\documentclass[conference]{IEEEtran}
\IEEEoverridecommandlockouts
\usepackage{subfigure}

\usepackage{pbalance}

\IEEEoverridecommandlockouts 
\usepackage[dvips]{graphicx}
\usepackage{algorithmic}
\usepackage{algorithm}
\usepackage[frozencache=true,cachedir=minted-cache]{minted} 
\usepackage[OT4,T1]{fontenc}
\usepackage[cmex10]{amsmath}
\usepackage{url}
\usepackage{multirow}
\usepackage{hyperref}

\title{Temporal Image Caption Retrieval Competition -- Description and Results}
\author{
\IEEEauthorblockN{Jakub Pokrywka, Piotr Wierzchoń, Kornel Weryszko, Krzysztof Jassem}
\IEEEauthorblockA{ Adam Mickiewicz University\\
Faculty of Mathematics and Computer Science,\\
Emails: jakub.pokrywka@amu.edu.pl, wierzch@amu.edu.pl \\
korwer@st.amu.edu.pl, jassem@amu.edu.pl 
}
}

\begin{document}

\maketitle              

\begin{abstract}
Multimodal models, which combine visual and textual information, have recently gained significant recognition. This paper addresses the multimodal challenge of Text-Image retrieval and introduces a novel task that extends the modalities to include temporal data. The Temporal Image Caption Retrieval Competition (TICRC) presented in this paper is based on the Chronicling America and Challenging America projects, which offer access to an extensive collection of digitized historic American newspapers spanning 274 years. In addition to the competition results, we provide an analysis of the delivered dataset and the process of its creation.
\end{abstract}

\section{Introduction}
Multimodal models are gaining great recognition, especially those combining image and text. A recent example is the image generation model, DALL·E 2 \cite{clip}. Tasks executed by such multimodal models usually consist of text-image retrieval, namely, either retrieving an image from its text description or retrieving a text caption for a given image. In this challenge, we introduce a task in the caption retrieval setup, additionally extending the model with temporal data.

Language models rarely utilize metadata, such as text domain, timestamp, or website URL. Additional temporal information may prove helpful when factual knowledge is required, and the facts rely on time (e.g., the answer to the question: “Who is the president of the U.S.A?” depends on the date). Temporal information may also be relevant in case of language semantic changes (e.g., the meaning of the word “gay” has shifted from “cheerful” to referring to homosexuality).

The presented task is based on the projects: Chronicling America \cite{chronicling} and Challenging America \cite{challam}. Chronicling America is an open database of over 16 million pages of digitized historic American newspapers covering 274 years. Challenging America is a set of temporal challenges based on the Chronicling America dataset.

The described competition was conducted using the Gonito platform \cite{gonito}, and its results are available at \url{https://gonito.csi.wmi.amu.edu.pl/challenge/cnlps-ticrc}. The competitions started on Feb 20, 2023, and ended on June 14, 2023. The training dataset was published in two batches (train and train2). Participants were allowed to use the delivered development dataset (dev) for training. The preliminary testing dataset (test-A) was available from the beginning of the competition. The final testing dataset (test-B) was released in the last two weeks of the competition. The golden truth for the testing datasets has not been made public. The Gonito platform is open to post-competition submissions.

\begin{figure}[h]
    \centering
    {
  \includegraphics[clip,width=1.0\columnwidth]{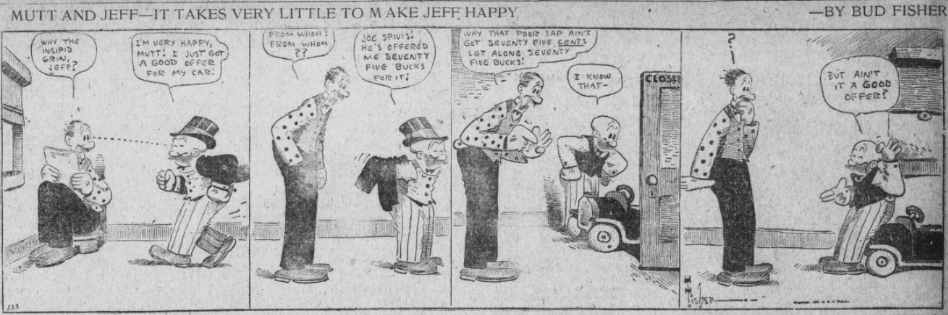}
           \caption{Sample picture with a caption above. This picture comes from a newspaper issued dated Jan 11, 1928.}
    \label{fig:samplepicture}
    }
\end{figure}

\section{Motivation}
From a linguistic and historical standpoint, Temporal Image Caption Retrieval (TICRC) holds significant value and brings various benefits. Firstly, TICRC facilitates the analysis of language evolution over time by associating image captions with specific temporal periods. Through this approach, researchers can investigate changes in vocabulary, grammar, and linguistic styles, thereby gaining insights into the adaptation and evolution of language across different historical contexts.

Secondly, TICRC contributes to the preservation and documentation of historical knowledge. Image captions accompanying visual content often contain valuable historical information. By leveraging TICRC, historians and researchers can effectively search and analyze these image captions, enabling a deeper understanding of specific historical periods, events, or cultural contexts. This process enhances the documentation of historical knowledge and enriches our comprehension of the past.

Furthermore, TICRC facilitates cross-referencing and integration of visual and textual sources. By associating image captions with specific temporal intervals, the competition makes it possible to establish connections between relevant textual documents, such as diaries, newspapers, or historical records. The interlinking of visual and textual data enhances contextualization and aids in interpreting and analyzing visual content from a historical perspective.

Moreover, TICRC offers valuable contextual information regarding the depicted scenes, individuals, or objects in images. By retrieving relevant captions based on temporal information, researchers gain a more comprehensive understanding of the context in which the images were captured. This contextualization further strengthens the interpretation and analysis of visual content within its historical framework.

In summary, Temporal Image Caption Retrieval enables the analysis of language evolution, enhances historical documentation and preservation, facilitates the integration of visual and textual sources, provides contextualization of visual content, and supports the study of cultural and societal changes over time.
\section{Related Work}

\subsection{Temporal language datasets and models}

Several textual benchmarks concerning the date of text publication have been published in recent years.  Challenging America \cite{challam} presents a set of three temporal tasks. Authors of \cite{temporalt5} introduce a temporal question answering task and dataset, in which the query's answer depends on a year, e.g., \textit{Who is the current president of the USA?}. Both benchmarks contain a baseline temporal language model trained on a text with a date timestamp prepended as text. In \cite{ireland}, the authors propose another text classification task, including temporal information. In addition to the timestamp in the textual form the model is also trained on temporal input embeddings. The authors of \cite{temporalattention} modify the transformer architecture, proposing a temporal attention component.

\subsection{Multimodal vision-language models}

Recently, the quality of vision-language models has improved greatly thanks to introducing models such as CLIP \cite{clip}, EVAL-CLIP \cite{evaclip}, ALIGN \cite{align}, BASIC \cite{basic}, LiT \cite{lit}, Flamingo \cite{flamingo}, or GPT-4 \cite{gpt4} and \cite{clipbaseline}. 

MS COCO \cite{mscoco} and Visual Genome \cite{visualgenome} are two large-scale, high-quality vision datasets annotated by humans. YFCC-100M \cite{yfcc} is an even larger dataset that contains user data collected from Flickr, not specifically designed for model training. Authors of  CC12M \cite{cc12m} and LAION-5B \cite{laion} apply cleaning procedures to adapt user data for the purpose of model training. The works mentioned did not prioritize the importance of temporal data.

\section{Task Definition}

The task here is to retrieve a relevant caption from a caption set for the given  picture from a newspaper and the newspaper's publication daily date. For each picture, only one caption is relevant.

The dataset is provided on the challenge GitHub repository \url{https://github.com/kubapok/cnlps-ticrc}.

Figure \ref{fig:samplepicture} presents an example source picture with a caption.

\subsection{Sample Data}
In this section, we provide sample data.  A picture and the publication date (in the YYYY-MM-DD format) of a given newspaper issue are given, as well as the collection of all captions for the given dataset type (train, train2, dev-0, test-A, or test-B). In the caption collection, a newline character is represented as \textbackslash n. The challenge participant is supposed to return the list of captions from the given dataset in descending probability order.

\textbf{Picture}: Figure \ref{fig:samplepicture}
\begin{figure}[h]
    \centering
    {
  \includegraphics[clip,width=1.0\columnwidth]{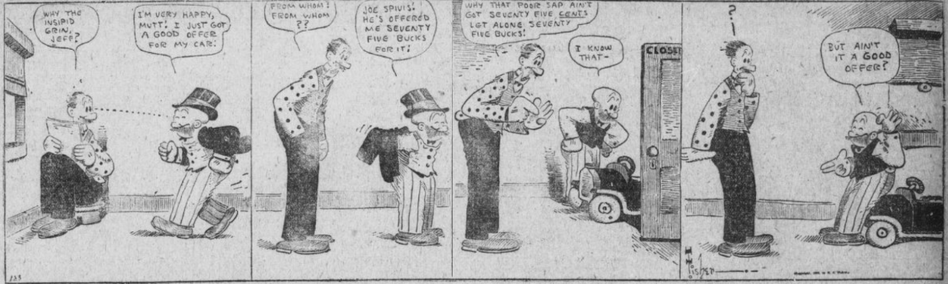}
           \caption{Sample input picture}
    \label{fig:samplepicture}
    }
\end{figure}

\textbf{Date timestamp}: 1928-01-11 

\textbf{Set of all possible captions}:
\begin{itemize}
    \item "China Dinner Sets."
    \item "MUTT AND JEFF –- IT TAKES VERY LITTLE TO MAKE JEFF HAPPY"
    \item "PARIS MILLINERY\texttt{\textbackslash}nfrom every Parisian modiste,\texttt{\textbackslash}nof note - embracing every \texttt{\textbackslash}nstyle tendency of the fall \texttt{\textbackslash}nand winter season \texttt{\textbackslash}nand \texttt{\textbackslash}n GOWNS COATS WRAPS \texttt{\textbackslash}nTAILORED SUITS AND \texttt{\textbackslash}nDRESSES"
    \item ...
\end{itemize}

\textbf{Correct Output}: "MUTT AND JEFF -–  IT TAKES VERY LITTLE TO MAKE JEFF HAPPY"

\begin{figure*}[t]
    \centering
\subfigure[1922-04-10 WINCHESTER\texttt{\textbackslash}nGolders'  Headquarters\texttt{\textbackslash}nFor Every Golfer]
    {
  \includegraphics[clip,width=0.6\columnwidth]{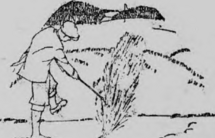}
  \label{fig:a1}
  }
\subfigure[1911-06-13 MICHELIN \texttt{\textbackslash}nInner Tubes \texttt{\textbackslash}nFor Michelin and all other Envelopes]
    {
  \includegraphics[clip,width=0.3\columnwidth]{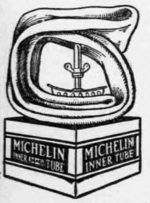}
  \label{fig:a2}
    }
\subfigure[1911-08-17 HERE'S the Shirt hit of the \texttt{\textbackslash}nseason]
    {
  \includegraphics[clip,width=0.7\columnwidth]{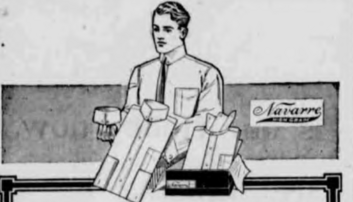}
  \label{fig:a3}
    }  
\subfigure[1913-02-12 DESKS AND OFFICE FURNITURE.]
    {
  \includegraphics[clip,width=0.6\columnwidth]{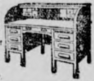}
  \label{fig:a4}
    }
\subfigure[1902-10-16 M. O'NEIL \& CO. \texttt{\textbackslash}nFurniture \texttt{\textbackslash}nAND...  \texttt{\textbackslash}nWe are now showing one of the \texttt{\textbackslash}nmost complete lines of \texttt{\textbackslash}nParlor \texttt{\textbackslash}nand \texttt{\textbackslash}nLibrary \texttt{\textbackslash}nTables]
    {
  \includegraphics[clip,width=0.52\columnwidth]{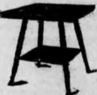}
  \label{fig:a5}
    }
\subfigure[1907-06-27 Guaranteed \texttt{\textbackslash}nPURE. \texttt{\textbackslash}nLEAD AND ZINC PAINTS. \texttt{\textbackslash}n"Made in BALTIMORE"]
    {
  \includegraphics[clip,width=0.45\columnwidth]{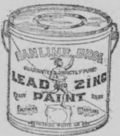}
  \label{fig:a6}
    }
\caption{Sample images from the training dataset with the corresponding date of publication caption. The images were not selectively chosen.}
    \label{fig:manysamples}
\end{figure*}

More examples are provided in Figure \ref{fig:manysamples}.
\subsection{Metric}

The metric for the competition is Mean Reciprocal Rank:

\begin{align*}
    \text{MRR}  &= \frac{1}{|Q|}\sum_{i=1}^{|Q|}\frac{1}{rank_i},
\end{align*}
where:
$|$Q$|$ ― number of queries, $rank_i$ ― rank position of the relevant document for the $i$-th query. The metric is implemented in the GEval evaluation tool \cite{geval} and available for offline use (details are provided on the competition page).


\section{Data Annotation Process}
\begin{figure}[h]
 \centering
  \includegraphics[clip,width=0.7\columnwidth]{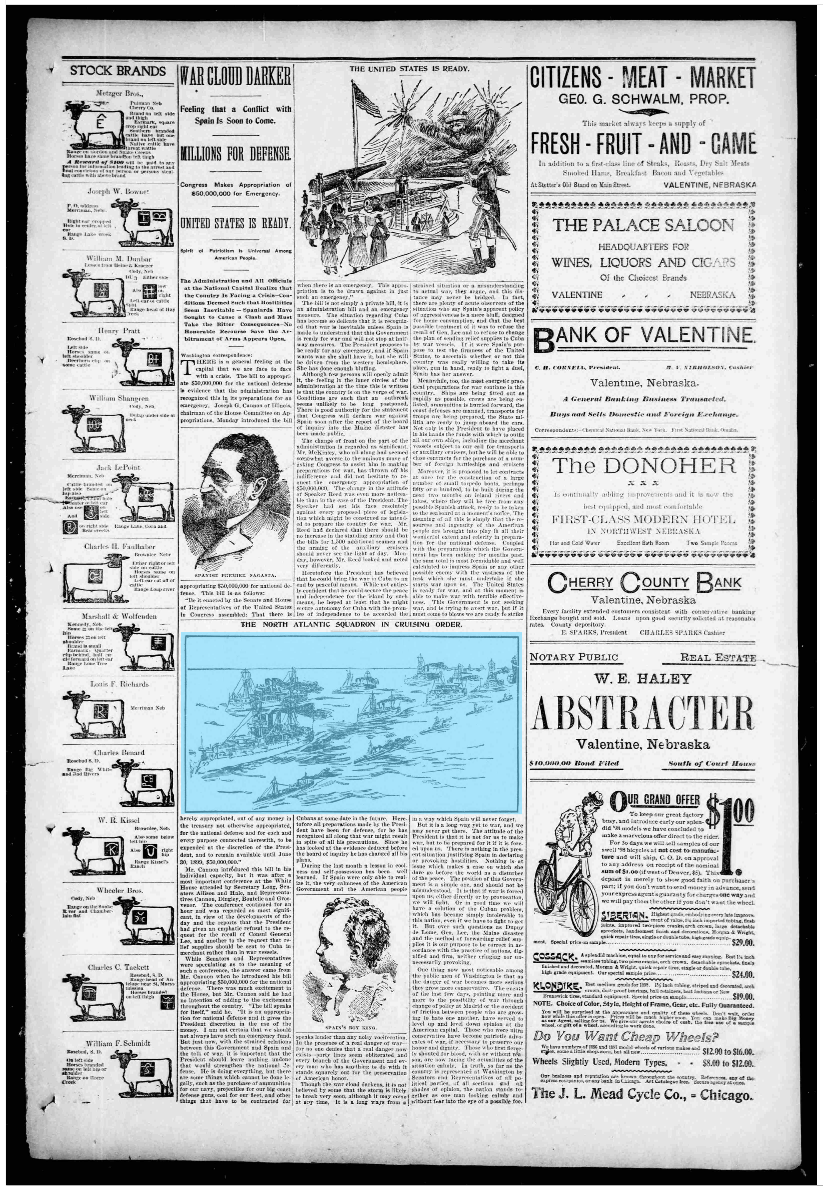}
  \label{fig:a6}
\caption{Picture selected on the whole page.} 
    \label{fig:calagazeta}
\end{figure}

The data was taken from the Challenging America project, according to the data processing rules provided there. The annotation was done manually in the Doccano \cite{doccano} system, which helped effective processing of annotation pairs: image and text. The annotation platform required the annotation of the entire newspaper pages. A sample page from which a picture was selected is presented in Figure \ref{fig:calagazeta}. The annotation of images was carried out according to given guidance rules divided into three aspects: Objects to be annotated (what to annotate), technical parameters of the image area (what technical requirements are imposed on annotated objects), and rules of text transcription (how to transcript caption texts).

These were the annotation guidance rules:
\paragraph{Objects to be annotated}
\begin{itemize}
  \item Images may be selected for annotation only if they occur along with the corresponding caption. 
  \item The caption text should be maximum a few sentences long. In case of longer captions, the annotator should select and mark the most relevant fragment of the caption.
  \item The caption text should -- at the discretion of the annotator -- be relevant to the image in content.
  \item The annotator should select at most one image per page.
  \item If the annotator has already encountered the same image on one of the previously annotated pages,  the image should not be annotated again.
  \item The annotator should minimize the number of portraits. 
\end{itemize}

\paragraph {Technical requirements for the image area (bbox)}
\begin{itemize}
\item The picture frame should encompass the image in its entirety (the picture should not be cut off).
\item The image frame should not cover more area than the image.
\item The frame must not cover the caption text.
\end{itemize}

\paragraph{Rules for text transcription}
\begin{itemize}
\item  The transcription should preserve the character size of the original
\item Punctuation and line-break characters should be preserved   as in the original.
\item Paragraph indentation in the text should be ignored. If the words are divided by a hyphen or line break, the original spelling (separated words) should be preserved.
\end{itemize}
The dataset was annotated mainly by one annotator, and his work took 70 hours.



\section{Data Analysis} 
The dataset comprises 3902 instances, each consisting of a picture, a caption, and a date timestamp. The pictures and corresponding captions were extracted from scans of newspapers dating back to 1853, which appends the element of fuzziness in image recognition to the challenge and makes the temporal aspect even more relevant (as the image quality depends on the publication date).

\subsection{Data Split}
Five datasets have been prepared for the competition -- two training sets (train, train2), a development set (dev-0), and two test sets (test-A, test-B). The final split ratio is illustrated in Table \ref{tab:data_split}. Precautions similar to those described in \cite{challam} have been taken to ensure that there is no detrimental overlap between the datasets.

\begin{table}[tbp]
\caption{Data split statistics}\label{tab:data_split}
\centering
\begin{tabular}{@{\vrule width0ptheight9pt\enspace}l|c|c|c}
\hfil\bf Type&\bf Name&\bf Instances& \bf Ratio\\
\multirow{2}{*}{Training} & train & 675 & \multirow{2}{*}{70.0}\\
& train2 & 2054 &\\\hline
Development & dev-0 & 646 & 16.6\\\hline
\multirow{2}{*}{Testing} & test-A & 92 & \multirow{2}{*}{13.4}\\
& test-B & 435 &\\
\end{tabular}
\end{table}

\subsection{Datasets Statistics}
For the sake of statistical analysis, the two testing datasets and the development dataset have been combined into one dataset, referred to as the testing dataset in this section. Similarly, the two training datasets have been combined into one.

\begin{figure}[tbp]
\centering
\includegraphics[width=0.75\hsize]{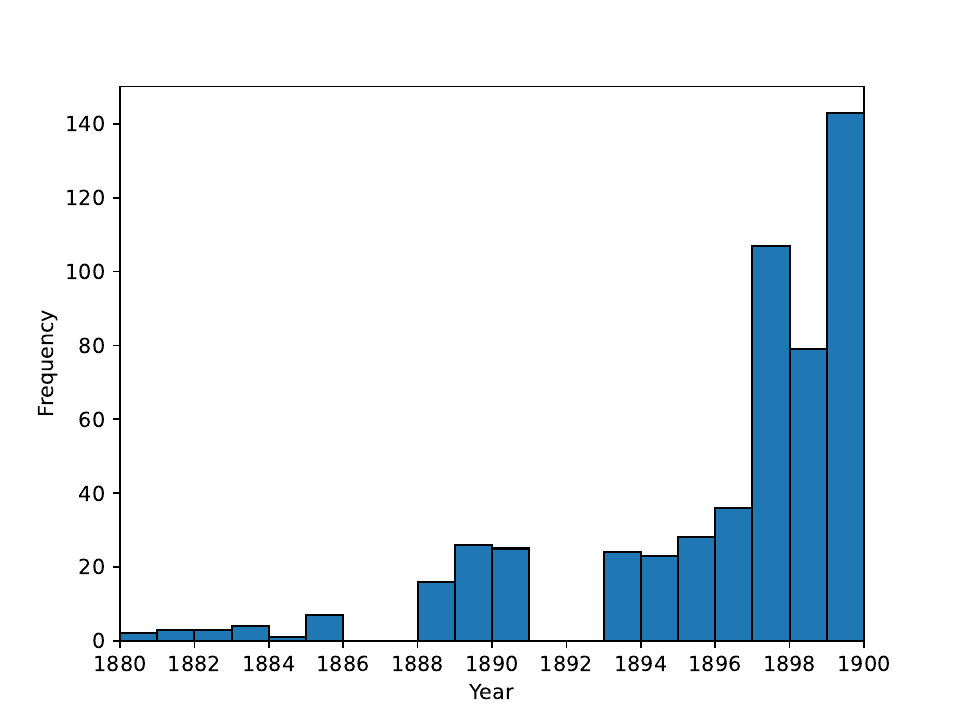}
\caption{Testing distribution over the years}
\label{fig:test_date_distribution}
\includegraphics[width=0.75\hsize]{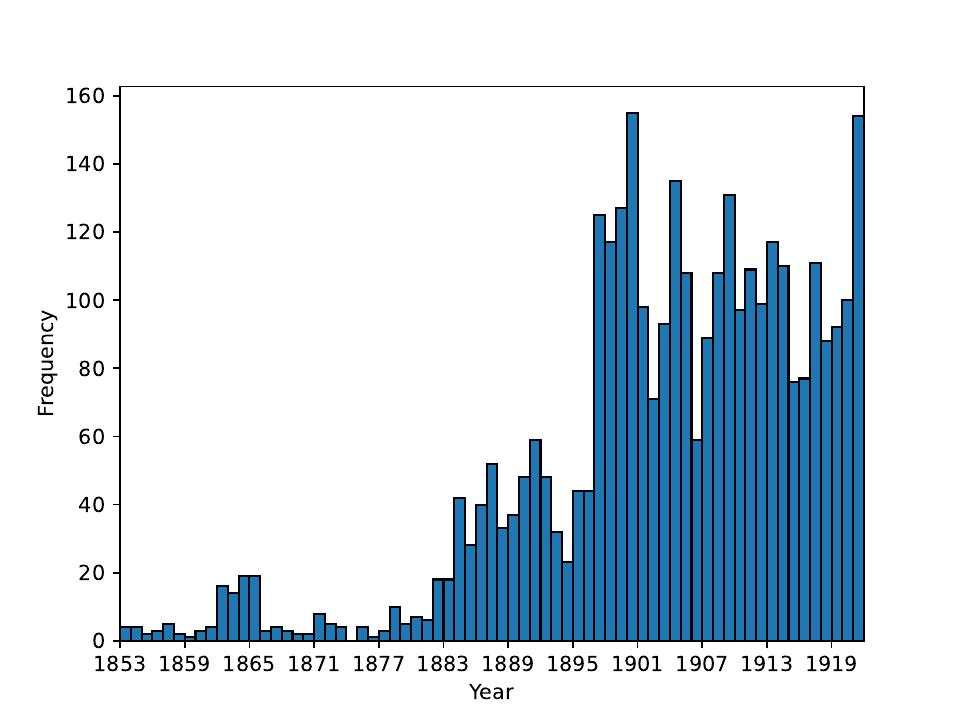}
\caption{Training distribution over the years}
\label{fig:train_date_distribution}
\end{figure}

Figures \ref{fig:test_date_distribution} and \ref{fig:train_date_distribution} provide insight into the temporal variance in the frequency distributions of the instances. Whereas both datasets are negatively skewed (as suggested by the mean $\approx 1895.82$ and median $=1897.0$ of the testing dataset and mean $\approx 1903.52$, median $=1905.0$ in the case of the training dataset), the latter covers a significantly greater period containing data points between 1853 and 1922. The testing dataset spans from 1880 to 1900. Moreover, the testing dataset's standard deviation $\approx 4.18$ is also less than $\frac{1}{3}$ of the training dataset's standard deviation $\approx 12.97$.

\begin{figure}[tbp]
\centering
\includegraphics[width=1.0\hsize]{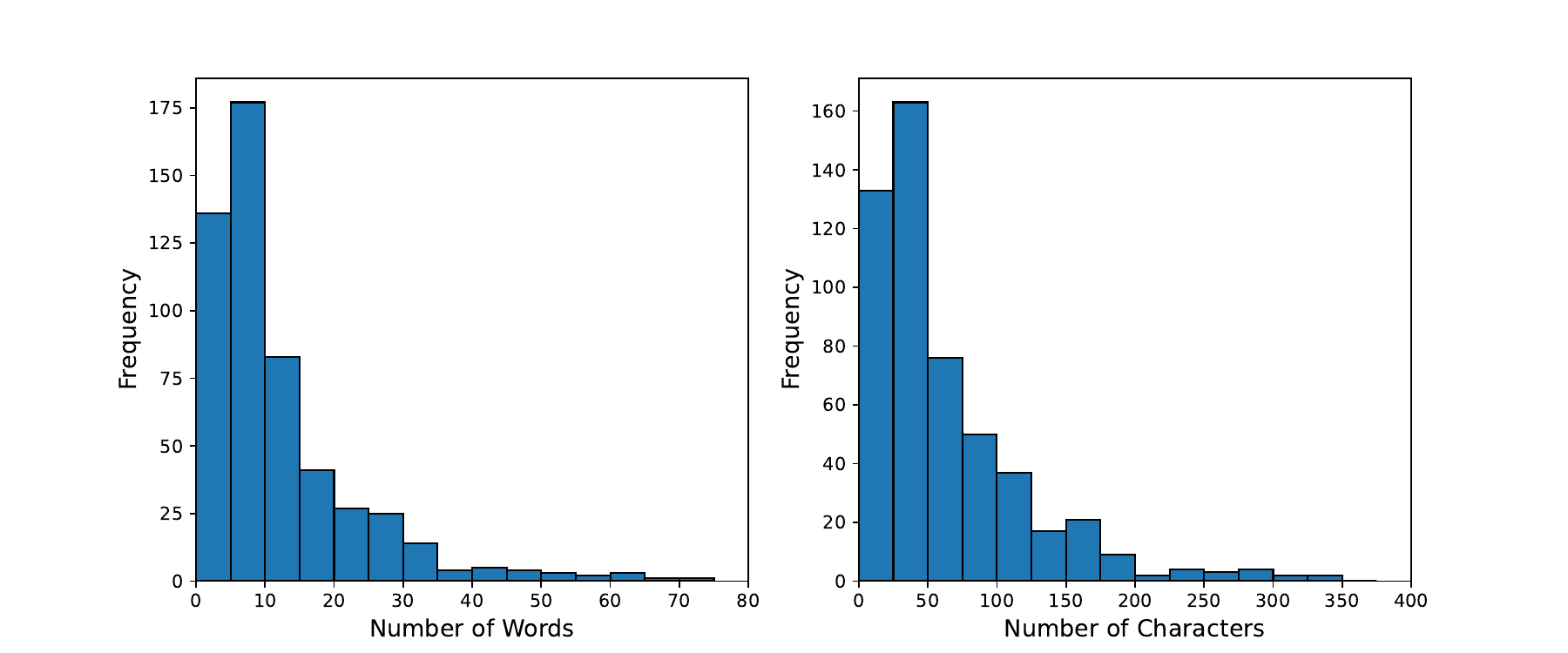}
\caption{Word and character per caption statistics in testing dataset}
\label{fig:test_caption_distributions}
\end{figure}

\begin{figure}[tbp]
\centering
\includegraphics[width=1.0\hsize]{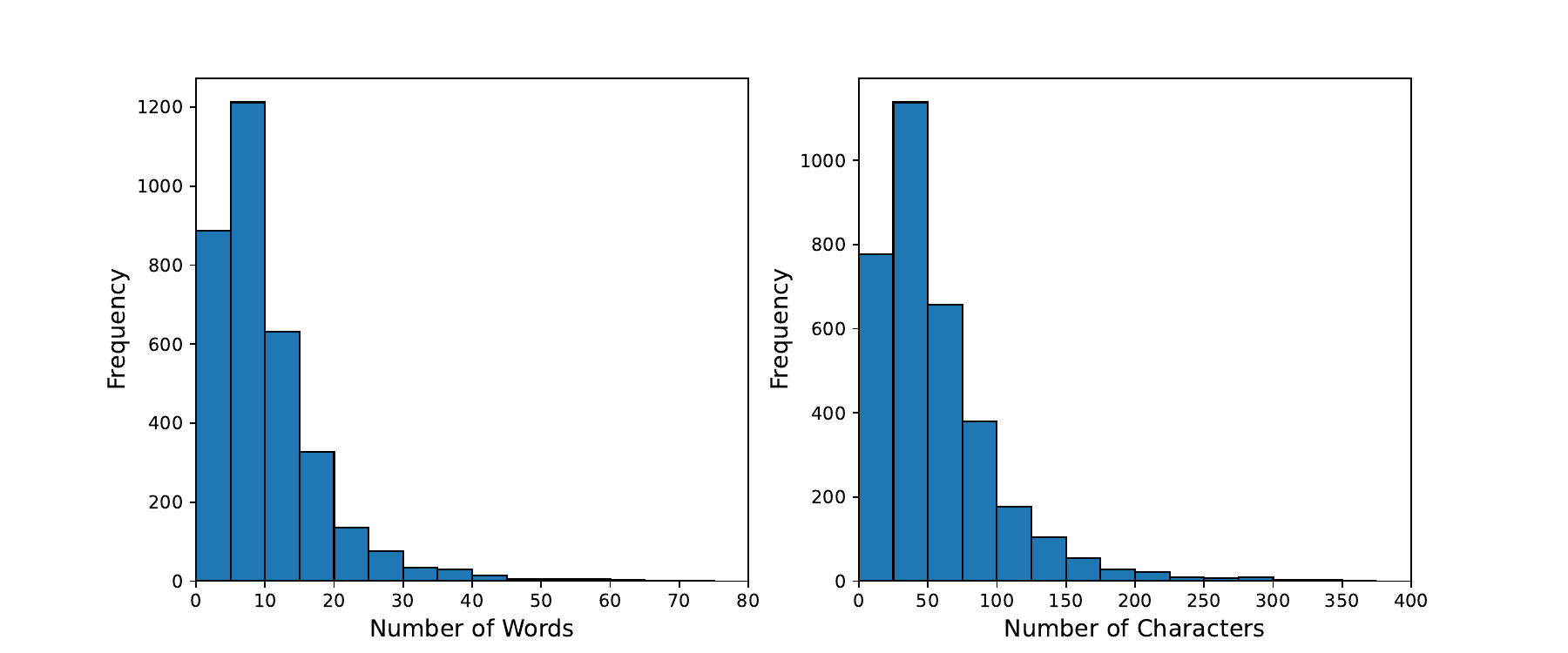}
\caption{Word and character per caption statistics in the training dataset}
\label{fig:train_caption_distributions}
\end{figure}

The captions are measured in the number of words and characters. The captions from the testing dataset captions tend to be longer, with mean $\approx 11.77$ and median $= 8.0$ words per caption and mean $\approx 66.79$, median $= 44.0$ characters per caption. The respective parameters for captions from the training dataset have the following values: mean $\approx 9.80$, median $= 7.0$ and mean $\approx 56.54$, median $= 43.0$. There is no significant difference in the corresponding frequency distributions, as can be seen in Figures \ref{fig:test_caption_distributions} and \ref{fig:train_caption_distributions}.

\section{Baselines}
The official competition baseline is included in the competition repository and relies on  the transformer model clip-ViT-B-32 \cite{clipbaseline} model without fine-tuning. The secondary baseline is the randomized caption order.

\section{Shared Task Results}
Five teams participated in the competition. Three solutions scored above the official competition baseline. The final results are provided in Table \ref{tab:results}.

\begin{table}[h]
    \centering
    \caption{Final competition results. The test-B dataset is used for winner determination, whereas the test-A dataset is only preliminary.} 
\begin{tabular}{c|c|c|c|c}
    place & submitter & test-A MRR & test-B MRR & submissions \\
    \hline

 1&      Kaszuba&                        0.6059  &   0.3444  &   6                     \\
 2 &     s478846&                        0.5529  &   0.33850  &   11                    \\
 3  &    Serba   &                       0.3506  &   0.2283  &   1                     \\
 -   &   \textbf{transformer baseline}&  0.2697  &   \textbf{0.1710}  &   -    \\
 4    &  Szyszko &                       0.0887 &    0.0621 &    1                     \\
 -   &   \textbf{random baseline}&       0.0513  & \textbf{0.0193} &   -    \\
 5     & s478855  &                      0.0514  &   0.0137  &   3                     \\

\end{tabular}
    \label{tab:results}
\end{table}

The competition's winner is Patryk Kaszuba, who was invited to prepare a report for publication in the conference proceedings and presentation at FedCSIS 2023. His solution is based on EVA02\_CLIP\_E\_psz14\_plus\_s9B model \cite{evaclip}. The model was used without fine-tuning to the competition dataset.

\section{Conclusions}

In this paper, we introduced a new benchmark for temporal image caption retrieval, called TRIC (Temporal Image Caption Retrieval). TRIC includes a three-modal (vision-language-time) dataset, divided into two train sets, two test sets and a development set. The proposed task consists in selecting a caption relevant  for a given image, from a given set. The temporal information is significant for the task as the data comprise scanned texts spanning the period of 274 years.

We organised the competition based on the benchmark. Five participants participated, with three of them scoring above the baseline. The benchmark is still open for further improvement of the obtained results.

We believe that TRIC will have a positive impact on the analysis of language evolution and support the study of cultural and societal changes over time.

%

\bibliography{bibliography}
\bibliographystyle{ieeetr}
\end{document}